\documentclass[runningheads]{llncs}
\usepackage{graphicx}
\usepackage{amsmath,amssymb} 
\usepackage{color}
\usepackage[dvipsnames]{xcolor}
\usepackage{comment}

\usepackage{booktabs}
\usepackage{hhline}
\usepackage{multirow}
\usepackage{ulem}

\begin{document}
\pagestyle{headings}
\mainmatter
\def\ECCVSubNumber{3720}  

\title{SPAN: Spatial Pyramid Attention Network for Image Manipulation Localization} 

\titlerunning{Spatial Pyramid Attention Network}
%
\author{Xuefeng Hu\inst{1} \and
Zhihan Zhang\inst{1}\and
Zhenye Jiang\inst{1}\and
Syomantak Chaudhuri \inst{2} \and
Zhenheng Yang \inst{3} \and
Ram Nevatia \inst{1}
}
\authorrunning{X. Hu, Z. Zhang et al}
%
\institute{
University of Southern California \\ \email{\{xuefengh,zhihanz,zhenyeji,nevatia\}@usc.edu}
\and Indian Institute of Technology, Bombay \\ \email{syomantak@iitb.ac.in}
\and Facebook AI \\ \email{zhenheny@fb.com}
}

\maketitle

\begin{abstract}
We present a novel framework, Spatial Pyramid Attention Network (SPAN) for detection and localization of multiple types of image manipulations. The proposed architecture efficiently and effectively models the relationship between image patches at multiple scales by constructing a pyramid of local self-attention blocks. The design includes a novel position projection to encode the spatial positions of the patches.  SPAN is trained on a generic, synthetic dataset but can also be fine tuned for specific datasets; The proposed method shows significant gains in performance on standard datasets over previous state-of-the-art methods. 
\end{abstract}

\section{Introduction}
\label{sec:intro}
Fast development of image manipulation techniques is allowing users to create modifications and compositions of images that look  ``authentic" at relatively low cost. Typical manipulation methods include splicing, copy-move, removal and various kinds of image enhancements  
to produce ``fake" or ``forged" images. We aim to localize   the manipulated regions of different tampering types in images.

\begin{figure}[!ht]
    \centering
    \includegraphics[width=\textwidth]{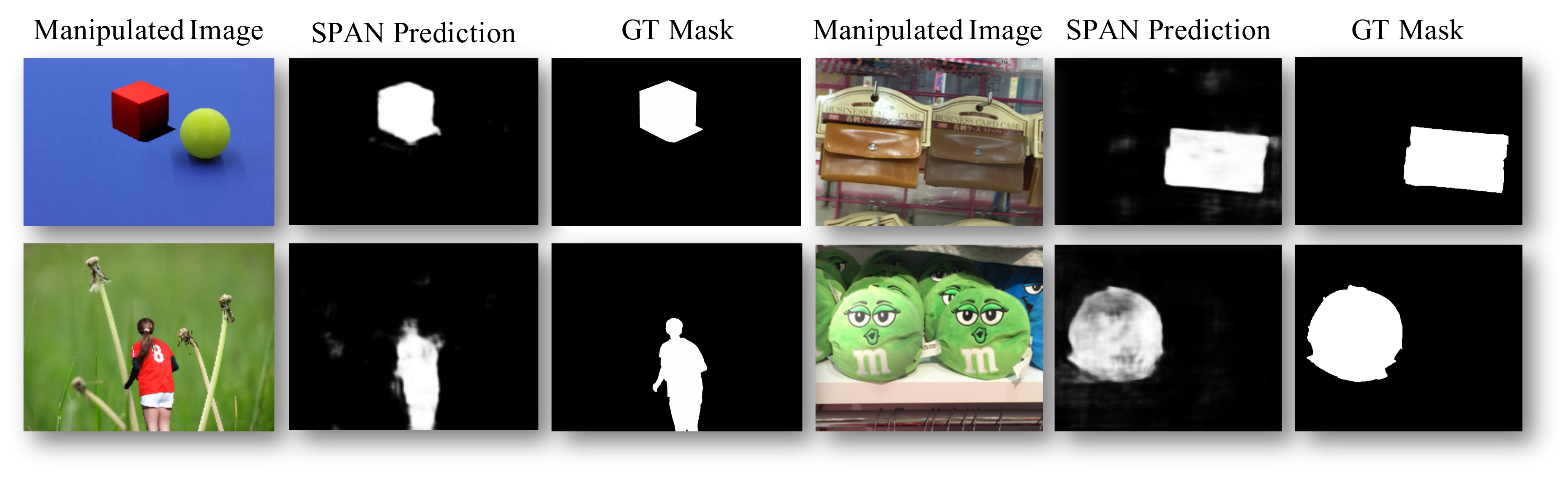}
    \caption{Examples of images manipulated by splicing (left) and copy-move (right) techniques, SPAN predictions and Ground-truth Masks.}
    \label{fig:demo1}
\end{figure}




There has been work on both detection and localization of manipulations in recent years. Among the methods that include localization, many  deal with only one or a few types of manipulations such as  splicing\cite{cozzolino2015splicebuster,patchconsist,nips2019}, copy-move\cite{cozzolino2015efficient,rao2016deep,wen2016coverage,wu2018busternet,wu2018image}, removal\cite{zhu2018deep}, and enhancement\cite{bayar2016deep,bayar2018constrained}. Some recent papers have  proposed more  general solutions that are not specific to  manipulation types; these include RGB-Noise (RGB-N) Net \cite{rgbn} and Manipulation Tracing Network (ManTra-Net) \cite{mantra}. The two methods differ in the granularity of their localization (bounding boxes in \cite{rgbn} \textit{vs} masks in \cite{mantra}). Our method is also designed for pixel-level masks predictions of multiple manipulation types.



A key assumption underlying forged region localization is that the  color, intensity or noise distributions of manipulated regions  are somehow different from those of the untampered ones. These distinctions are typically not transparent to a human observer but the expectation is that they can be learned by a machine. 
Ability to capture and model the relationship between tampered and authentic regions is crucial for this. 
RGB-N \cite{rgbn} proposes a Faster R-CNN \cite{ren2015faster} based method to detect tampered region; hence, its predictions are limited to the rectangular box. It also requires fine-tuning to adapt to a new dataset.
ManTra-Net \cite{mantra} combines the task of image manipulation type classification and localization into a unified framework and achieves comparable results to RGB-N but without fine-tuning on the target evaluation data.  ManTra-Net makes pixel-level predictions. However, its localization module only models the ``vertical'' relationship of same points on different scales of the feature map but does not model the spatial relations between image patches. We propose to model both ``vertical'' relationship using multi-scale propagation and spatial relationship using a self-attention module in the Spatial Pyramid Attention Network (SPAN).



SPAN is composed of three blocks: a feature extractor, a spatial pyramid  attention block and a decision module. We adopt the pre-trained feature extractor provided by \cite{mantra} as our feature extractor. To build the relationship between different image patches, we construct a hierarchical structure of self-attention layers with the features as input.

The hierarchical structure is designed to first calculate local self-attention blocks, and the local information is propagated through a pyramid hierarchy. This enables an efficient and multi-scale modeling of neighbors. Inspired by \cite{tansformer}, within each self-attention layer, we calculate a new representation for each pixel conditioned on its relationship to its neighborhood pixels. To better capture the spatial information of each neighbor in the self-attention block, we also introduce a positional projection which is more suitable for our localization task, instead of the original positional embedding methods used for machine translation task in \cite{tansformer}.
The decision module, which is composed of a few 2D convolutional layers, is applied on top of the output from pyramid spatial attention propagation module to predict the localization mask.



We conducted comprehensive experiments on five different popular benchmarks and compared with previous works on pixel-level manipulation localization of different tampering types. Our proposed SPAN model outperforms current state-of-the-art (SoTA) methods ManTra-Net \cite{mantra} and RGB-N \cite{rgbn} with the same setting on almost every benchmark (e.g., 11.21\% improvement over ManTra-Net on Columbia, without fine-tuning; 9.5\% improvement over RGB-N on Coverage, with fine-tuning).

In summary we make three contributions in this work: (1) we  designed a novel Spatial Pyramid Attention Network architecture that efficiently and explicitly compares patches through the local self-attention block on multiple scales; (2) we introduced positional projection in the self-attention mechanism, instead of classic positional embedding used in text processing;  (3) We show that the ability to compare and model relationship between image patches at different scales results in higher accuracy compared to state-of-the-art methods.

\section{Related Work}

  \textbf{Manipulation Detection and Localization:}   
    Most of existing methods  detect and localization a specific type of manipulation only, such as copy-move\cite{cozzolino2015efficient,rao2016deep,wen2016coverage}, removal\cite{zhu2018deep}, enhancement\cite{bayar2016deep,bayar2018constrained}, and  splicing\cite{cozzolino2015splicebuster,patchconsist,nips2019,salloum2018image}. Among these types, splice detection is the most studied topic.  For this task, MFCN\cite{salloum2018image} has proposed a Multi-task Fully Convolutional Network \cite{long2015fully} with enhancement on edge detection, jointly trained towards both edge detection and tampered region detection. MAG\cite{nips2019} proposed Generative-Adversarial based pipeline to train its model with augmented retouched samples, based on a new splicing dataset.

    
    The types of manipulation, however, may not always be known in advance and a single image may contain   multiple types of manipulations. Therefore, it is important to have systems that detect general types of manipulations. Some recent work \cite{mantra,rgbn,bappy2019hybrid,bappy2019hybrid} has addressed this problem, and shown success in building models that have robustness to detection of  multiple manipulation techniques. J-LSTM\cite{bappy2017exploiting} is an LSTM based patch comparison methods that finds tempered regions by detecting edges between the tempered patches and authentic patches. H-LSTM\cite{bappy2019hybrid}  improved this method by introducing a separate encoder-decoder structure to refine the predicted mask, and a Hilbert Curve route to process the image patches, to establish better context information. 
    Both methods operate on fixed size image patches, which might cause failure when the tempered region size does not follow this assumption. 
    
    RGB-N\cite{rgbn} proposes a two-stream Faster R-CNN network\cite{ren2015faster}, which combines the regular RGB-based Faster R-CNN with a parallel module that works on the noise information  generated by the Steganalysis Rich Model (SRM) \cite{fridrich2012rich}. RGB-N is pre-trained on a synthetic tampering dataset generated base on MS-COCO \cite{lin2014microsoft}. Due to the R-CNN architecture, RGB-N is limited to localizing to a rectangular box whereas real objects are not necessarily rectangular. RGB-N also requires fine tuning on specific datasets to achieve high accuracy.

    ManTra-Net\cite{mantra} has proposed a system that jointly learns the Manipulation-Tracing feature for both image manipulation classification and forgery localization. ManTra-Net is composed of a VGG\cite{simonyan2014very} based feature extractor and an LSTM\cite{hochreiter1997long} based detection module. The feature extractor is trained towards  385 types of image manipulation, then it is trained together with the detection module on a synthetic dataset for multi-type manipulation detection. While \cite{rgbn,bappy2017exploiting,bappy2019hybrid}  require fine-tuning on different manipulation distributions to achieve state-of-the-art performance, ManTra-Net achieves comparable results without any training on the domain datasets. 
\subsubsection{Attention Mechanism}

    Attention mechanism, first introduced by \cite{bahdanau2014neural}, has brought tremendous benefit to the many fields, such as  Machine Translation \cite{bahdanau2014neural,tansformer}, Image Captioning \cite{xu2015show}, Image Question and Answering \cite{yang2016stacked} and Object Detection \cite{ba2014multiple,hu2018squeeze,li2019selective}. Attention mechanism helps the neural network build input-aware connections to focus more on meaningful entities, such as words or regions, by replacing the classic learnable fixed weights with input dependent weights.  
    
    Self-Attention mechanism proposed in \cite{tansformer} calculates mutual attention between a group of input, has succeed in capturing long term word relationships within the same sentence, and has been dominating Machine Translation in the recent two years. Image transformer \cite{imagetf} has made an attempt to bring the self-attention mechanism to image generation, by adopting an encoder-decoder structure based on pixel-wise self-attention. The Non-local Neuron Network\cite{nonlocal} uses self-attention mechanism to model non-local relationship between pixel features, and  achieves an improvement in activity recognition, image classification and object detection. However, the non-local blocks are of $O(N^4)$ complexity in both times and space, for $N\times N$ size input, which implies that it could only be applied when the spatial resolution is reduced. For tasks addressed in \cite{nonlocal}, good results are obtained by applying the non-local layer to reduced resolution maps; however, for manipulation detection, it may be harmful to reduce spatial resolution  as the small differences between pixels that are important in tracing forgery regions,  may be lost during the pooling operations. 

\section{Method}



We describe our proposed Spatial Pyramid Attention Network (SPAN) in this section. We first provide an overview of the framework, then the details of each module and lastly how the whole framework can be trained.


\subsection{Overview}\label{overview}
 SPAN is composed of three blocks: feature extractor, pyramid spatial attention propagation and decision module. We adopt the pre-trained feature extractor as proposed in \cite{mantra}. The feature extraction network is trained on synthesized data generated based on  images sampled from the \textit{Dresden Image Database} \cite{gloe2010dresden}, using classification supervision against 385-type image manipulation data. The feature extractor adopts the Wider \& Deeper VGG Network \cite{simonyan2014very,zagoruyko2016wide} as the backbone architecture, along with Bayer \cite{bayar2016deep} and SRM \cite{fridrich2012rich,zhou2017two} Convolutional Layers to extract rich features from both visual artifacts and noise patterns. 

\begin{figure}[t]
    \centering
    \includegraphics[width=1.01\linewidth]{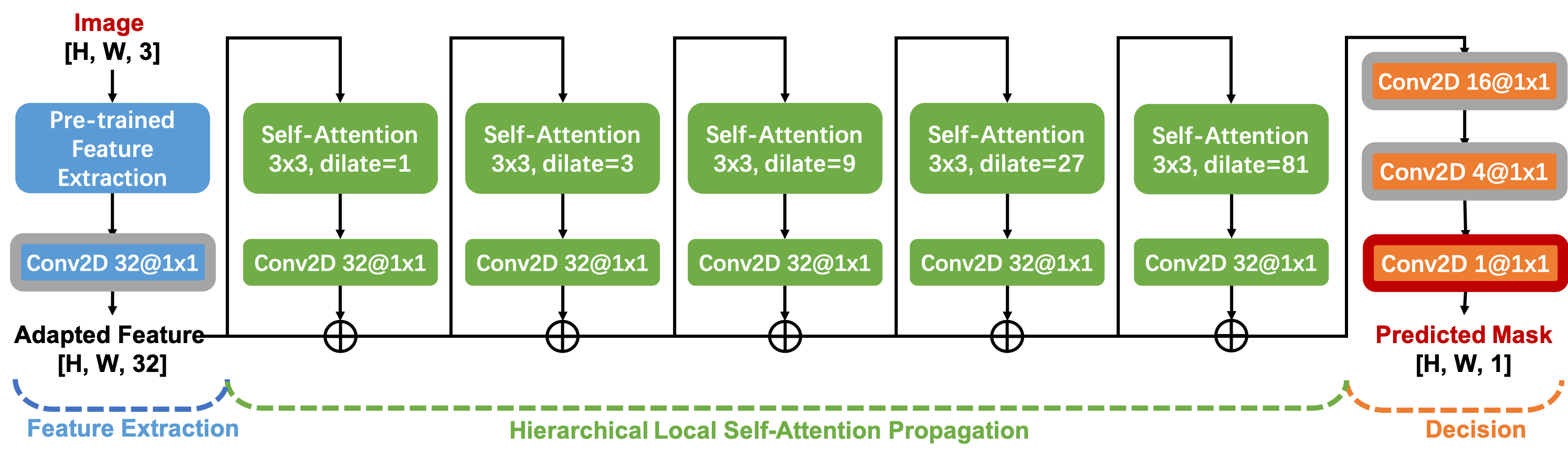}
    \caption{Overview of the  Spatial Pyramid Attention Network. There are three blocks in SPAN framework: feature extraction (blue), pyramid spatial attention propagation (green) and decision module (orange).}
    \label{fig:overall}
\end{figure}

On top of the embeddings extracted using the pre-trained feature extractor, we apply a convolution layer to adapt the feature to feed into the spatial attention module.
The pyramid spatial attention propagation module is designed to establish the spatial relationship on multiple scales of the pixel representation.
Five layers of the local self-attention blocks are recursively applied to extract information from neighboring pixels or patches. To preserve the details from multi-scale layers, we add the input back to the output after each self-attention block, through the residual link which is proposed in \cite{he2016deep}.


To generate the final mask output, 2D convolutional blocks are applied. A Sigmoid activation is applied to predict the soft tampering mask. 





\subsection{Local Self-Attention Block}
\label{self-attn}
    Consider an image feature tensor $X$ of size $H\times W\times D$, where $H$ and $W$ are the spatial dimensions, and $D$ is the feature dimension. For simplicity, we use $X_{i,j}$ to represent pixel at $i$-th row and $j$-th column, where each $X_{i,j}\in \mathbb{R}^D$.
    We calculate the Local Self-Attention value, $LSA(X_{i,j}|X,N,t)$, for each position $X_{i,j}$, by looking at it and its $(2N+1) \times (2N+1)$ neighborhood, dilated by distance $t$. For example, when $N=1$, the $3\times3$ neighborhood  looks like:
    $$
    	\left[
        	\begin{matrix}
        	X_{i-t,j-t} & X_{i-t,j} & X_{i-t,j+t}\\
        	X_{i,j-t} & X_{i,j} & X_{i,j+t}\\
            X_{i+t,j-t} & X_{i+t,j} & X_{i+t,j+t}
            \end{matrix}
        \right]
    $$
    
    For simplicity, we rename the $(2N+1)^2$ neighbors linearly from top-left to bottom-right, as $\{Y_1, Y_2, ... , Y_{(2N+1)^2}\}$. Then, the LSA over $X_{i,j}$ is calculated as 
    \begin{equation}
    \label{selfattneq}
        LSA(X_{i,j}|X,N,t) = \frac{1}{C}\sum_{l=1}^{(2N+1)^2} \exp{\left(\frac{\langle M^kY_l, M^qX_{i,j} \rangle}{\sqrt{D}}\right)} M^vY_l
    \end{equation}
    where $\langle,\rangle$ represents the inner product, $C=\sum_{l=1}^{(2N+1)^2} \exp{\left(\frac{\langle M^kY_l, M^qX_{i,j} \rangle}{\sqrt{D}}\right)}$ is the soft-max normalization factor, and $M^k, M^q, M^v \in \mathbb{R}^{D\times D}$ are learnable weights.
    
    Equation \ref{selfattneq} can be rewritten as 
    
    \begin{equation}
        LSA(X_{i,j}|X,N,t)^\top = SoftMax\left(\frac{(M^qX_{i,j})^\top M^kP_{i,j}}{\sqrt{D}}\right)(M^vP_{i,j})^\top
    \end{equation}
    where $P_{i,j}=[Y_1,...,Y_{(2N+1)^2}]\in \mathbb{R}^{D\times (2N+1)^2}$, and $M^kP_{i,j}, M^vP_{i,j}, M^qX_{i,j}$ correspond to the terminology of Keys, Values and Query in attention mechanism literature \cite{tansformer,nonlocal,bahdanau2014neural}.
    
    Relationships between neighborhood $\{Y_1, Y_2, ... , Y_{(2N+1)^2}\}$ and pixel $X_{i,j}$ can be explicitly modeled by the quadratic form $\langle M^kY_l, M^qX_{i,j}\rangle$, and used to help build the feature for manipulation localization. Such information cannot be easily extracted by any convolution network with similar number of parameters, as convolution blocks only adds pixels together, rather than building their mutual relationship.
    \label{position}
      \begin{figure}[t]
        \centering
        \includegraphics[width=\linewidth]{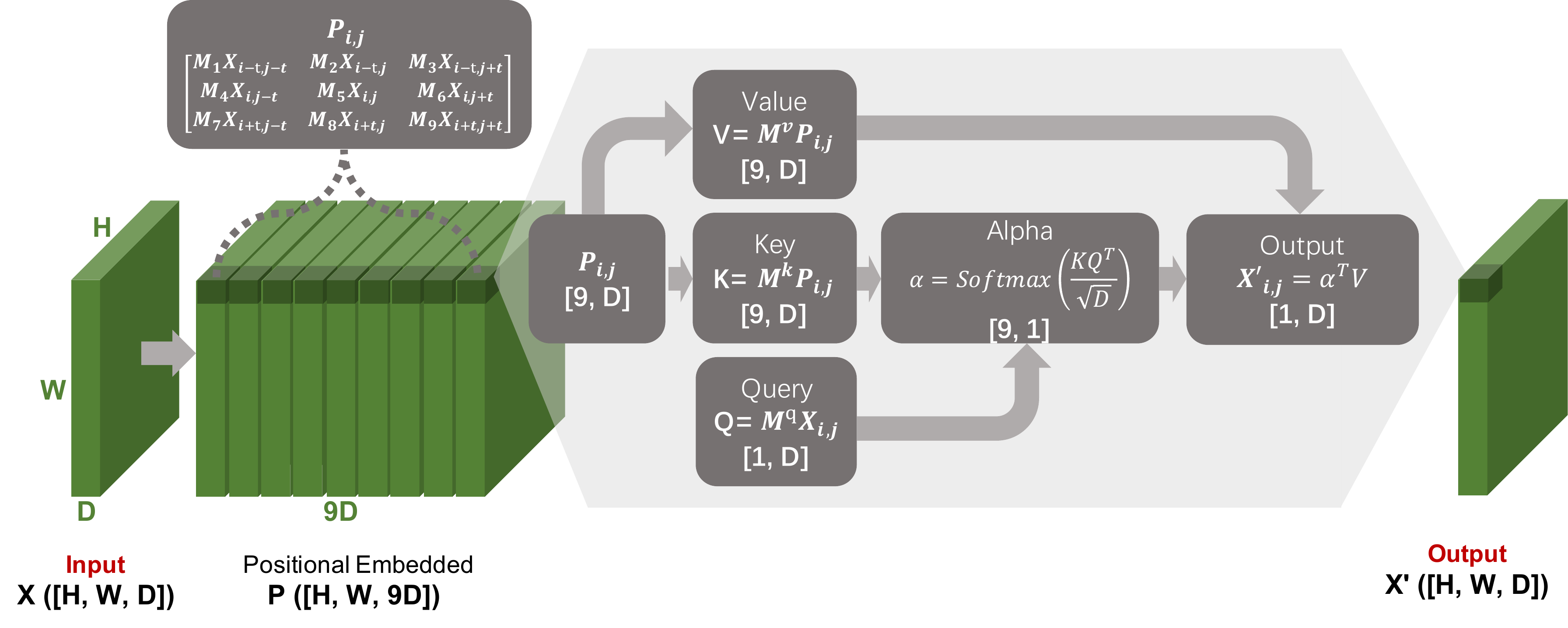}
        \caption{Self-Attention Block: We perform self-attention mechanism over each target pixel $X_{i,j}$ and its local neighborhood $P_{i,j}$. With learnable projections $M^q, M^k$ and $M^v$, we prepare Keys and Values from $P_{i,j}$ and Query from $X_{i,j}$ for the attention mechanism.}
        \label{fig:selfattn}
    \end{figure}
\subsection{Positional Projection} 
    As shown in the Transformer work \cite{tansformer}, positional embedding is essential in self-attention calculation, to enable the model to tell the difference between inputs with different temporal order. 
    In our case, it is also important that the model know the relative spatial relationship between the neighbor pixel and query pixel. 
    In machine translation models \cite{tansformer}, positional embeddings are learnable weights that directly add to inputs at each input position. This is reasonable for language tasks, because the word embedding space is typically trained to support vector addition and subtraction  corresponding to linguistic meaning. However, as our embedding space does not follow these requirements, we propose to use learnable matrix projections $\{M_l\}_{1\le l\le (2N+1)^2}$ to represent the $(2N+1)^2$ possible relative spatial relationships.
    With positional projection, the LSA block now becomes:
    \begin{equation}
    \label{self-attn-final}
        LSA(X_{i,j}|X,N,t) = \frac{1}{C}\sum_{l=1}^{(2N+1)^2} \exp{\left(\frac{\langle M^kM_lY_l, M^qX_{i,j} \rangle }{\sqrt{D}}\right)} M^vY_l
    \end{equation}
    with soft-max normalization factor $C=\sum_{l=1}^{(2N+1)^2} \exp{\left(\frac{\langle M^kM_lY_l, M^qX_{i,j} \rangle }{\sqrt{D}}\right)}$.

    In Figure \ref{fig:selfattn} we illustrate  a local self-attention block  when $N=1$. Given an image tensor $X$, we first prepare the neighborhood $P_{i,j}$ with 9 positional projected neighbors for each pixel $X_{i,j}$. Note that for edge and corner pixels, the neighborhood sizes are 6 and 4 respectively. The neighbors are then projected into Keys, Values and Query through matrix projections $M^k,M^v$ and $M^q$. Finally, Keys, Values and Query are assembled into output $X_{i,j}'$ (Equation \ref{self-attn-final}).
\subsection{Pyramid Propagation}
    
    The local self-attention blocks compare pixels and their neighbors with limited distances $Nt$. However, images might have large tampered regions, which require comparison between pixels far away from each other. Instead of simply enlarging $N$ to have better pixel coverage, we iteratively apply our local self-attention blocks to propagate the information in a pyramid structure.
    
        \label{pyramid}
    \begin{figure}[t]
        \centering
        \includegraphics[width=0.8\linewidth]{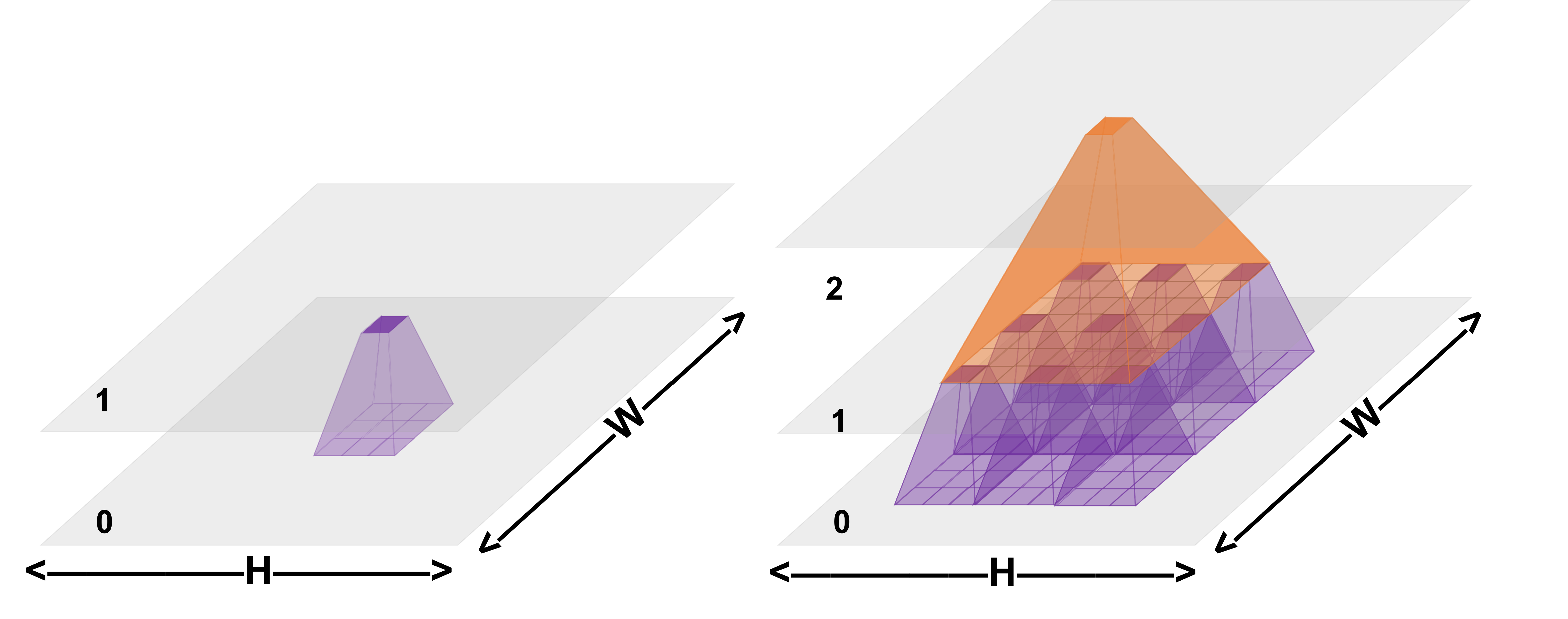}
        \caption{Local Attention Pyramid Propagation: Going through self-attention blocks with proper dilation distances (e.g. dilation distance of 1 in the purple blocks and 3 in the orange blocks), information from each pixel location is propagated in a pyramid structure (e.g. each pixel on level-2 encodes information from 9 pixels on level-1 and 81 pixels on level-0).} %
        \label{fig:layers}
    \end{figure}

    As shown in Figure \ref{fig:layers}, the purple and orange stacks represent local self-attention blocks, each with $N=1$, $t=1$ and $N=1$, $t=3$. Through the self-attention block, each pixel on layer 1 represents 9 pixels from layer 0; each pixel on layer 2 represents 9 pixels from layer 0, and therefore 81 pixels from layer 1. With properly set dilation distances, pixels from the top of $h$-layer local self-attention structure can reach to $(2N+1)^{2h}$ pixels from the bottom layers. 
    
    There are two benefits of the pyramidal design: 1) We can use a small value for $N$, which makes each self-attention block efficient to compute; 2) in upper layers, pixels not only represent themselves, but encode information from their local regions. Therefore, by comparing those pixels, the self-attention blocks compare information from two different patches.
    
    \subsubsection{Analysis of block size}
        As demonstrated in Section \ref{pyramid}, a $h$-layer $M\times M$ self-attention block structure with dilation distances $\{1,M,...,M^{h-1})\}$ helps the output pixels related to most pixels from the input, where $M=2N+1$ is the neighborhood size of each self-attention block. Each pixel in the final layer covers a $M^h\times M^h$ region on the input features. Therefore, to cover a $S\times S$ size image, $h=O(\log_M S)$ layers would be needed. As $S^2$ self-attention calculation is needed at each layer, and each self-attention block is of $O(M^2)$ complexity in both time and memory, the total complexity with respect to $M$ is $O(S^2M^2\log_M S)$, which reaches to its minimum with $M=2N+1=3$, among possible integer values. 
        
        The  $3\times 3$ blocks not only offer the largest coverage under the same memory and time budget, but also offer comparisons over more scales. In our  SPAN implementation, we adopt a 5-layer, $3\times 3$ self-attention blocks structure, with dilation distances $1,3,9,27, 81$, as demonstrated in Figure \ref{fig:overall}. The proposed structure compares and models the relationship between neighbor regions at five different scales, of 3, 9, 27, 81 and 243.
   
\subsection{Framework Training}
    To train SPAN, we freeze the feature extractor and train the following parts in an end-to-end fashion, supervised by binary ground-truth mask with 1 labels tampered pixels and 0 labels authentic pixels. We adopt Binary Cross-Entropy (BCE) as our loss function:
    \begin{equation*}
        \label{loss}
        \displaystyle
        Loss(X^{pred}, B) = 
            \frac{1}{HW}\sum_{i=1}^{H}\sum_{j=1}^{W}  \left(-B_{i,j}\log X^{pred}_{i,j}-\left(1-B_{i,j}\right)\log \left(1-X^{pred}_{i,j}\right)\right)
    \end{equation*}
    where $X^{pred}$ is the output predicted mask, and $B$ is the binary ground-truth mask. $B_{i,j}$ and $X^{pred}_{i,j}$ represent the corresponding pixel value at location $i,j$.

\section{Experiments}
    In this section, we describe experiments on five different datasets to explore the effectiveness of SPAN model. These datasets include (1) base training dataset as proposed in \cite{mantra}; (2) NIST16 \cite{NIST16}; (3) Columbia \cite{ng2009columbia}; (4) Coverage \cite{wen2016coverage}; (5) CASIA \cite{dong2013casia}. We follow the evaluation protocols as in \cite{mantra,rgbn,nips2019}. We compare with other state-of-the-art (SoTA) methods under the setup with and without finetuning for specific datasets. We compare only with methods that attempt to detect general manipulations and not tuned to a specific manipulation type, such as splicing\cite{salloum2018image,nips2019}.  
\subsection{Datasets}
\label{datasets}
    Following the practices in \cite{mantra}, we used the synthetic dataset proposed in \cite{mantra} for training and validation. Four other popular benchmarks for manipulation detection are used for fine-tuning and evaluation. 
    \begin{itemize}
        \item \textbf{Synthetic Dataset\cite{mantra}}  is composed of four subsets: the removal and enhancements datasets from \cite{mantra}, the splicing dataset from \cite{wu2017deep} and the copy-move dataset from \cite{wu2018busternet}. All samples from the synthetic dataset are of size $224\times 224$. 
        \item \textbf{NIST16\cite{NIST16}} contains 584 samples with  binary ground-truth masks. Samples from NIST16 are manipulated by one of the types from Splicing, Copy-Move and Removal, and are post-processed to hide visible traces. A 404:160 training-testing split is provided by \cite{rgbn}.
        \item \textbf{Columbia\cite{ng2009columbia}} is a Splicing based datset, containing 180 images, with provided edge masks. We  transform the edge masks into binary region masks, with positive tampered pixels and negative authentic pixels. All images in Columbia are used for testing only.
        \item \textbf{Coverage\cite{wen2016coverage}} is a Copy-Move based dataset, containing only 100 samples, with provided binary ground-truth masks, with post-processing to remove the visible traces of manipulation. A 75:25 training-testing split is provided by \cite{rgbn}.
        \item \textbf{CASIA\cite{dong2013casia}} is composed of CASIAv1 and CASIAv2 splits. CASIAv2 contains 5123 images, CASIAv1 contains 921 images; both with provided binary ground-truth masks. Samples from both subsets are manipulated by either Splicing or Copy-Move operations. Image enhancement techniques including filtering and blurring are applied to the samples for post-processing. According to \cite{rgbn}, CASIAv2 is the train split, and CASIAv1 is the test split.
    \end{itemize}
    
      \setlength{\tabcolsep}{3pt}
    \begin{table}[t]
         \caption{Pixel-level localization performance under the metrics of AUC and $F_1$ comparison on validation set  (SynVal) of the pre-training data and four different benchmarks. We could not find the model that ManTra-Net reported in the paper. Hence, we also report the performance of ManTra-Net's default GitHub model where we share the same feature extractor.}
           \centering
      \fontsize{9}{11}\selectfont
        \def\arraystretch{1.1}
        \begin{tabular}{lccccc}
        \specialrule{.2em}{.1em}{.1em}
              & SynVal & Columbia & Coverage & CASIA & NIST16\\
        \cline{2-6}
              & $F_1$   & AUC   & AUC   & AUC   & AUC \\
        \midrule
        ManTra-Net \cite{mantra} & 48.39 & 82.4   & 81.9  & \textbf{81.7} & 79.5  \\
        ManTra-Net (GitHub) & - & 77.95   & 76.87  & 79.96 & 78.05  \\
        \midrule
        SPAN  & \textbf{81.45} & \textbf{93.61}& \textbf{92.22} & 79.72 & \textbf{83.95}  \\
        \bottomrule
        \end{tabular}%
        \label{tab:withmantra}
    \end{table}%

\subsection{Implementation Details}
\label{implmentation}
    As discussed in Section \ref{pyramid} and demonstrated in Figure \ref{fig:overall}, we adopt the 5-layer $3\times 3$ self-attention block structure, with dilation distance $1,3,9,27,81$. We use the residual link to preserve information from each level, and use the new proposed positional projection to capture the spatial relationships between neighbors. We report the performance using this setting, unless otherwise specified; ablation on different model choices is given in Section \ref{abs}. 
    
    \label{basetrain}
    To train SPAN, we set batch size to 4 with 1000 batches per epoch. The batches are uniformly sampled from the synthetic dataset described above. The models are optimized by the Adam optimizer \cite{kingma2014adam} with initial learning $10^{-4}$ without decay. The validation loss, precision, recall and $F_1$ are evaluated at each epoch. Learning rate is halved if the validation loss fails to decrease per 10 epochs, until it reaches $10^{-7}$. The training is stopped early if the validation loss fails to decrease for 30 epochs. The best model is picked according to the validation loss.

    Following the setting from \cite{rgbn,bappy2017exploiting,wu2018busternet}, we  also fine-tune our model on the training splits from the four standard benchmarks for some of our experiments, after it is trained on the synthetic data. For NIST16 and Coverage, we use the exact same split provided by RGB-N \cite{rgbn}; for CASIA, we use CASIAv2 as training and CASIAv1 as testing, as stated in RGB-N \cite{rgbn}. 
    
    Stopping criteria for fine-tuning are set as following:
    For NIST16 and Converage, we use cross validation instead of using a fixed validation set, due to the small training split sizes. For CASIA, we randomly sampled 300 images from its training split to construct a validation split. The model with the best validation AUC scores are picked and used for evaluation on test set.
    
    Note that the pre-trained feature extractor achieves its peak performance on original size images, as described in \cite{mantra}. However, a fixed input size is required for the detection module, for fine-tuning and generalization. Therefore, during fine-tuning and inference stage, we extract the features from original sized image first, and then resize the feature tensor into fixed resolution of 224x224, for optimal performance. Total run time for a 256x384 image is around 0.11 second. 
        
\subsection{ Evaluation and Comparison}
\label{abs}
    We  evaluate and compare SPAN with current SoTA methods \cite{mantra,rgbn,bappy2017exploiting,bappy2019hybrid,krawetz2007picture,mahdian2009using,ferrara2012image} on the four benchmarks under different setups: (1) pre-training only; (2) pre-training + fine-tuning. We also explored the effectiveness of each proposed module by conducting ablation studies on the four standard benchmarks.  We use pixel-level Area Under the Receiver Operating Characteristic Curve (AUC) and $F_1$ score for comparison against the state-of-the-art general type forgery localization methods, according to \cite{rgbn} and \cite{mantra}. 
    

    

    \textbf{Pre-training only.} We compare the performances of SPAN and ManTra-Net \cite{mantra} under the same setup: both models are trained on synthetic dataset as in \cite{mantra} and evaluated on validation split of the synthetic dataset (SynVal) and four different datasets. To fairly compare with \cite{mantra}, all images in these four datasets are used as test data. As presented in Table \ref{tab:withmantra}, SPAN outperforms ManTra-Net by a large margin on SynVal. With the same feature extraction backbone, our SPAN is more effective of modeling the spatial relationship among different patches and thus generates better localization accuracy. Comparing with ManTra-Net \cite{mantra} on other four datasets, SPAN also shows better performance under the metric of AUC, demonstrating that our proposed model has good capability of generalizing to other datasets without extra adaptation. 
    
    SPAN achieves performance gains of over 10\% in AUC compared to ManTra-Net on Columbia (\cite{ng2009columbia}) and Coverage \cite{wen2016coverage}. The performance boost on CASIA \cite{dong2013casia} is not as significant. There are two possible reasons: 1) Tampered regions in Columbia and Coverage have more varied scales than those in CASIA dataset 
    and our SPAN also benefits from our more effective multi-scale modeling
    ; 2) CASIA samples have lower average resolution ($256\times 384$) compare to the other datasets (NIST: $2448\times 3264$; Columbia: $666\times 1002$; Coverage: $209 \times 586$). As we show in Table \ref{robustness}, the performance gap between SPAN and ManTra-Net decreases when test images are resized to lower resolution. 
    
    

   
        \setlength{\tabcolsep}{2pt}
        \begin{table}[t]
         \caption{Pixel-level localization performance under the metrics of AUC and $F_1$ comparison on validation set of the pre-training data and four different benchmarks. For NIST16, Coverage and CASIA, all models are fine-tuned on the corresponding training splits unless specifically stated. *We found there is an overlap of images between training and test data of NIST16. }
          \centering
          \fontsize{10}{12}\selectfont
            \begin{tabular}{llcccccccc}
            \specialrule{.2em}{.1em}{.1em}
            \multirow{2}{*}{} & \multirow{2}{*}{Supervision} & \multicolumn{2}{c}{Columbia} & \multicolumn{2}{c}{Coverage} & \multicolumn{2}{c}{CASIA} & \multicolumn{2}{c}{NIST16*} \\
        \cmidrule(lr){3-4}  \cmidrule(lr){5-6} \cmidrule(lr){7-8} \cmidrule(lr){9-10}   &    & AUC   & $F_1$    & AUC   & $F_1$    & AUC   & $F_1$    & AUC   & $F_1$ \\
            \midrule
            ELA\cite{krawetz2007picture}& unsupervised  & 58.1 & 47.0 & 58.3 & 22.2 & 61.3 & 21.4 & 42.9 & 23.6 \\
            NOI1\cite{mahdian2009using}& unsupervised & 54.6 & 57.4 & 58.7 & 26.9 & 61.2 & 26.3 & 48.7 & 28.5\\ 
            CFA1\cite{ferrara2012image}& unsupervised & 72.0 & 46.7 & 48.5 & 19.0 & 52.2 & 20.7 & 50.1 & 17.4 \\
            \midrule
            J-LSTM\cite{bappy2017exploiting}& fine-tuned &  -    & -     & 61.4  &  -    & -     & -     & 76.4  & - \\
            H-LSTM\cite{bappy2019hybrid}& fine-tuned &  -    & -     & 71.2  &  -    & -     & -     & 79.4  & - \\
            RGB-N\cite{rgbn}& fine-tuned & 85.8  & 69.7  & 81.7  & 43.7  & 79.5  & \textbf{40.8}  & 93.7* & \textbf{72.2*} \\
            \midrule
            SPAN & pre-training  & \textbf{93.6} & \textbf{81.5} & 91.2  & 53.5 & 81.4 & 33.6 & 83.6 & 29.0 \\
            SPAN & fine-tuned &  -    & -     & \textbf{93.7} & \textbf{55.8} & \textbf{83.8} & 38.2 & \textbf{96.1*} & 58.2* \\
            \bottomrule
            \end{tabular}%
          \label{tab:withrgbn}%
        \end{table}%

        
        
        \textbf{Pre-training + fine-tuning.} We now compare SPAN with other SoTA methods \cite{bappy2017exploiting,bappy2019hybrid,rgbn} under the fine-tuning setup. We also report scores of traditional unsupervised signal analysis models (ELA\cite{krawetz2007picture}, NOI1\cite{mahdian2009using} and CFA1\cite{ferrara2012image}) here as they are also evaluated over the testing splits rather than over the whole dataset.
        For fair comparison, we follow the same practices as in \cite{rgbn}: (1)directly evaluate on Columbia dataset; (2) finetune our model on Coverage and CASIA training split and evaluate on test split; (3) finetune the model on training split of NIST16 provided by \cite{rgbn} and evaluate on test split. The results as shown in Table \ref{tab:withrgbn} demonstrate that SPAN without fine-tuning already outperformed RGB-N and other methods by a large margin on Columbia and Coverage, further proving that our our spatial attention module has the strong capability of generalization. With fine-tuning, the performances on all four datasets further improve.
        
        J-LSTM and H-LSTM also make predictions by comparing image patches. Two possible reasons our method achieves better performance  are: 1)  J-LSTM and H-LSTM  look at patches at a single scale only, which might  limit them from detecting tempered regions that are very large or very small;  2) J-LSTM treats patches independently and adopts an LSTM structure to process patches linearly, which might lose track of the spatial and context information of those patches; H-LSTM attempts to alleviate this problem by taking in patches along a specifically designed Hilbert-Curve route, but it could still be hard for a linear LSTM to explicitly model spatial information. SPAN considers patches with its context and neighbor pixels, and has the ability to model spatial relationship though the positional projection.
        
        There is not a large performance gain on CASIA dataset compared to RGB-N \cite{rgbn}, which is pre-trained on object detection dataset (MS-COCO dataset \cite{lin2014microsoft}). One observation on CASIA dataset is that compared to Columbia and Coverage, where there is a combination of both tampered objects and random tampered region, most tampering on CASIA images occur on objects. We put NIST16 numbers in the last column as we observed there are visually almost identical images in both training and testing splits follow the same protocol in \cite{rgbn}. 
\textbf{Manipulation type analysis.} In Table \ref{tab:type} we present SPAN's performance on different manipulation types when evaluated on NIST16 \cite{NIST16} dataset. For comparison, we also generate the per-class results by directly evaluating the model provided in ManTra-Net GitHub \footnote{https://github.com/ISICV/ManTraNet}. Without fine-tuning on NIST16 (comparing the first tow rows), our SPAN model performs consistently better than ManTra-Net on all three manipulation types, demonstrating that our proposed spatial attention model is effective agnostic to tampering types. SPAN results can be further improved with adaptation onto this specific dataset. 
        \setlength{\tabcolsep}{4pt}
        \begin{table}[t]
         \caption{Pixel-level AUC and $F_1$ comparison on multiple manipulation types evaluated NIST16\cite{NIST16} dataset.}
          \centering
          \fontsize{10}{11}\selectfont
            \begin{tabular}{lcccccc}
            \specialrule{.2em}{.1em}{.1em}
            \multirow{2}[4]{*}{} & \multicolumn{2}{c}{Splicing} &  \multicolumn{2}{c}{Removal} & \multicolumn{2}{c}{Copy-Move}  \\
        \cmidrule(lr){2-3}
        \cmidrule(lr){4-5}
        \cmidrule(lr){6-7}
        & AUC   & $F_1$    & AUC   & $F_1$    & AUC   & $F_1$  \\
            \midrule
            ManTra-Net \cite{mantra} (GitHub) & 85.89 & 	38.56 & 	65.52 & 	14.86 & 	79.84 & 	15.03  \\
            \midrule
            SPAN (pre-training)  & 90.27 &	42.66 &	77.15 &	15.73 &	82.82 &	13.81  \\
            SPAN (fine-tuned) &  \textbf{99.15} &	\textbf{82.94} &	\textbf{90.95} &	\textbf{49.95} &	\textbf{90.94} &	\textbf{40.54} \\
            \bottomrule
            \end{tabular}%
          \label{tab:type}%
        \end{table}%

                \setlength{\tabcolsep}{5pt}
        \begin{table}[t]
        \caption{Comparison of SPAN variants evaluated with $F_1$ metric on SynVal, Columbia, Coverage, CASIA and NIST16}
          \centering
          \fontsize{10}{12}\selectfont
            \begin{tabular}{lccccc}
            \specialrule{.2em}{.1em}{.1em}
            F1 & SynVal & Columbia & Coverage & CASIA & NIST16\\
        \midrule
            SPAN (Res)          & 79.36 & 68.76 & 44.53 & 31.56 & 27.92\\
             SPAN (Res$+$PE)    & 79.72 & 68.38 & 49.13 & 26.01 & 28.05\\
            SPAN (LSTM$+$PP)    & 78.99 & 80.56 & 49.09 & 29.78 & 27.70 \\
            SPAN (Res$+$PP) & \textbf{81.76} & \textbf{81.45} & \textbf{53.47} & \textbf{33.63} & \textbf{28.99}\\
            \bottomrule
            \end{tabular}%
          \label{tab:ablation}%
        \end{table}%
        
        
        
      \begin{table}[!htbp]
        \caption{Robustness Analysis of SPAN over NIST16 and Columbia Dataset}
        \centering
        \begin{tabular}{lcccc}
        \specialrule{.2em}{.1em}{.1em}
                             & \multicolumn{2}{c}{NIST16} & \multicolumn{2}{c}{Columbia} \\ 
        \cmidrule(lr){2-3}  \cmidrule(lr){4-5} Pixel AUC               & SPAN & ManTra-Net & SPAN & Mantra-Net \\ \hline
        No manipulation               & 83.95                         & 78.05                         & 93.6                              & 77.95                             \\ 
        Resize (0.78x)                & 83.24                         & 77.43                         & 89.99                             & 71.66                             \\ 
        Resize (0.25x)                & 80.32                         & 75.52                         & 69.08                             & 68.64                             \\ \hline
        No manipulation               & 83.95                         & 78.05                         & 93.6                              & 77.95                             \\ 
        GaussianBlur (kernal size=3)  & 83.1                          & 77.46                         & 78.97                             & 67.72                             \\ 
        GaussianBlur (kernal size=15) & 79.15                         & 74.55                         & 67.7                              & 62.88                             \\ \hline
        No manipulation               & 83.95                         & 78.05                         & 93.6                              & 77.95                             \\ 
        GaussianNoise (sigma=3)       & 75.17                         & 67.41                         & 75.11                             & 68.22                             \\ 
        GaussianNoise (sigma=15)      & 67.28                         & 58.55                         & 65.8                              & 54.97                             \\ \hline
        No manipulation               & 83.95                         & 78.05                         & 93.6                              & 77.95                             \\ 
        JPEGCompress (quality=100)    & 83.59                         & 77.91                         & 93.32                             & 75                                \\ 
        JPEGCompress (quality=50)     & 80.68                         & 74.38                         & 74.62                             & 59.37                             \\ \bottomrule
        \end{tabular}
      
        \label{robustness}
        \end{table}

        \textbf{Ablation studies.} We explored how much each proposed component contributes to the final performance. We explore: (1) how to combine the outputs from different layers of multi-scale attention module (convolution LSTM (LSTM) \cite{xingjian2015convolutional} and Residual link (Res)); (2) how to model self-attention (position projection (PP) and position embedding (PE)). Besides the variants in the two modules, the number of self-attention hierarchy is set to 5 and $N=1$. The comparison of different variants is presented in Table \ref{tab:ablation}. Comparing the first two rows, Residual link performs better than LSTM on fusing the multi-scale attention feature. One possible explanation is Residual link is easier to optimize compared to the more complex LSTM; with a strong feature encoded from our attention module, SPAN with Residual link  converges better. 
        We also compare the performances of SPAN using two types of position modeling: position embedding as in \cite{tansformer} and our proposed position projection. We compare three variants Res+PP, Res+PE and Res only in Table. \ref{tab:ablation}. Under the setup of SPAN model, Res and Res+PE perform similarly across different datasets.
        Replacing PE with PP as the positional modeling achieves significant performance improvement on all datasets.
        
   

\textbf{Robustness} 
We conducted experiments to explore the robustness of SPAN to various manipulation types in Table \ref{robustness}. 
To produce modified samples, we apply standard OpenCV built-in functions \texttt{AREAResize}, \texttt{GaussianBlur}, \texttt{GaussianNoise}, \texttt{JPEGCompress} on NIST16 and Columbia. 
SPAN demonstrates more robust performance to compression but is more sensitive to resizing. 

\begin{figure}[t]
        \centering
        \includegraphics[width=\textwidth]{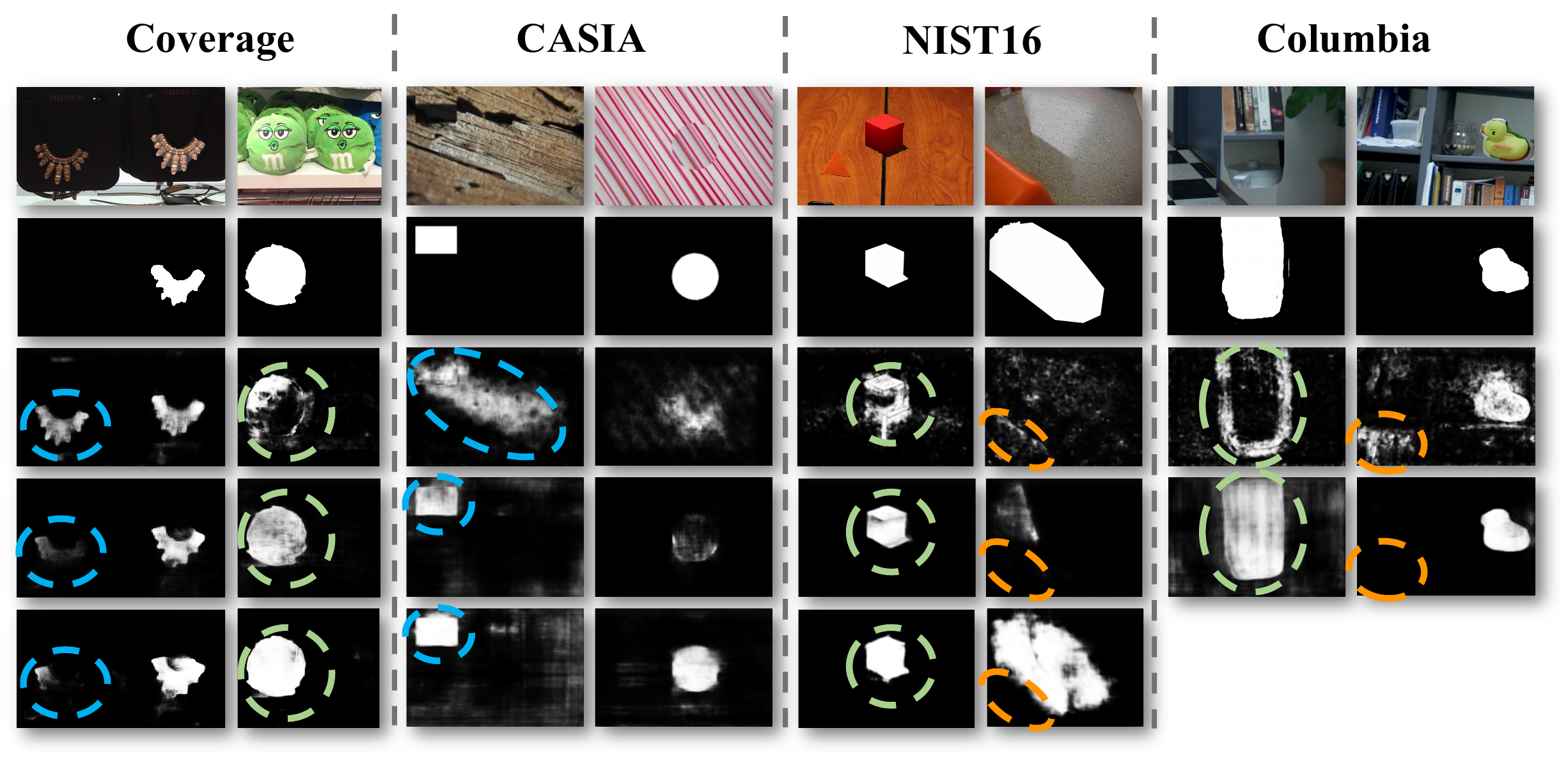}
        \caption{Comparison of SPAN prediction results on Coverage, CASIA, Columbia NIST16 datasets, with ManTra-Net predictions. From top to bottom: Manipulated Image, Ground-truth mask, ManTra-Net prediction, SPAN prediction and fine-tuned SPAN prediction. There is no fine-tuning result on Columbia dataset as there is no training split in Columbia.}
        \label{fig:visual1}
    \end{figure}

\textbf{Qualitative Results.} Figure \ref{fig:visual1} shows some SPAN and MantraNet \cite{mantra} results.
The examples show that SPAN produces better predictions  compared to Man-Tra Net in three commonly observed cases: 1) interior of tempered regions (green circles); 2) the correct manipulated object (blue circles) and 3)  noisy false negative predictions  (orange circles). 
 


\section{Conclusion}

We presented a Spatial Pyramid Attention Network (SPAN) that models the relationships between  patches on multiple scales through a pyramid structure of  local self-attention blocks, to detect and localize multiple image manipulation types with or without fine tuning.
SPAN outperforms the state-of-the-art models \cite{rgbn,mantra}.The  method is both accurate and robust in general type manipulation detection and localization, indicating that modeling patch relationship at different scales help capture the essential information in manipulation localization. However, SPAN may be less effective with lower image resolution.

\section*{Acknowledgement}
    
This work is based on research sponsored by the Defense
Advanced Research Projects Agency under agreement number
FA8750-16-2-0204. The U.S. Government is authorized
to reproduce and distribute reprints for governmental purposes
notwithstanding any copyright notation thereon. The
views and conclusions contained herein are those of the authors
and should not be interpreted as necessarily representing
the official policies or endorsements, either expressed
or implied, of the Defense Advanced Research Projects
Agency or the U.S. Government. We thank Arka Sadhu for valuable discussions and suggestions.

\clearpage

\bibliographystyle{splncs04}
\bibliography{egbib}
\end{document}


\pagestyle{headings}
\mainmatter
\def\ECCVSubNumber{3720}  

\title{Supplementary Material for \\ SPAN: Spatial Pyramid Attention Network for Image Manipulation Localization} 

\titlerunning{Spatial Pyramid Attention Network}
%
\author{Xuefeng Hu\inst{1} \and
Zhihan Zhang\inst{1}\and
Zhenye Jiang\inst{1}\and
Syomantak Chaudhuri \inst{2} \and
Zhenheng Yang \inst{3} \and
Ram Nevatia \inst{1}
}
\authorrunning{X. Hu, Z. Zhang et al}
%
\institute{
University of Southern California \\ \email{\{xuefeng,zhihanz,zhenyeji,nevatia\}@usc.edu}
\and Indian Institute of Technology, Bombay \\ \email{syomantak@iitb.ac.in}
\and Facebook AI \\ \email{zhenheny@fb.com}
}

\maketitle

\section*{Dataset Characteristics}
\begin{figure}
        \centering
        \includegraphics[width=\linewidth]{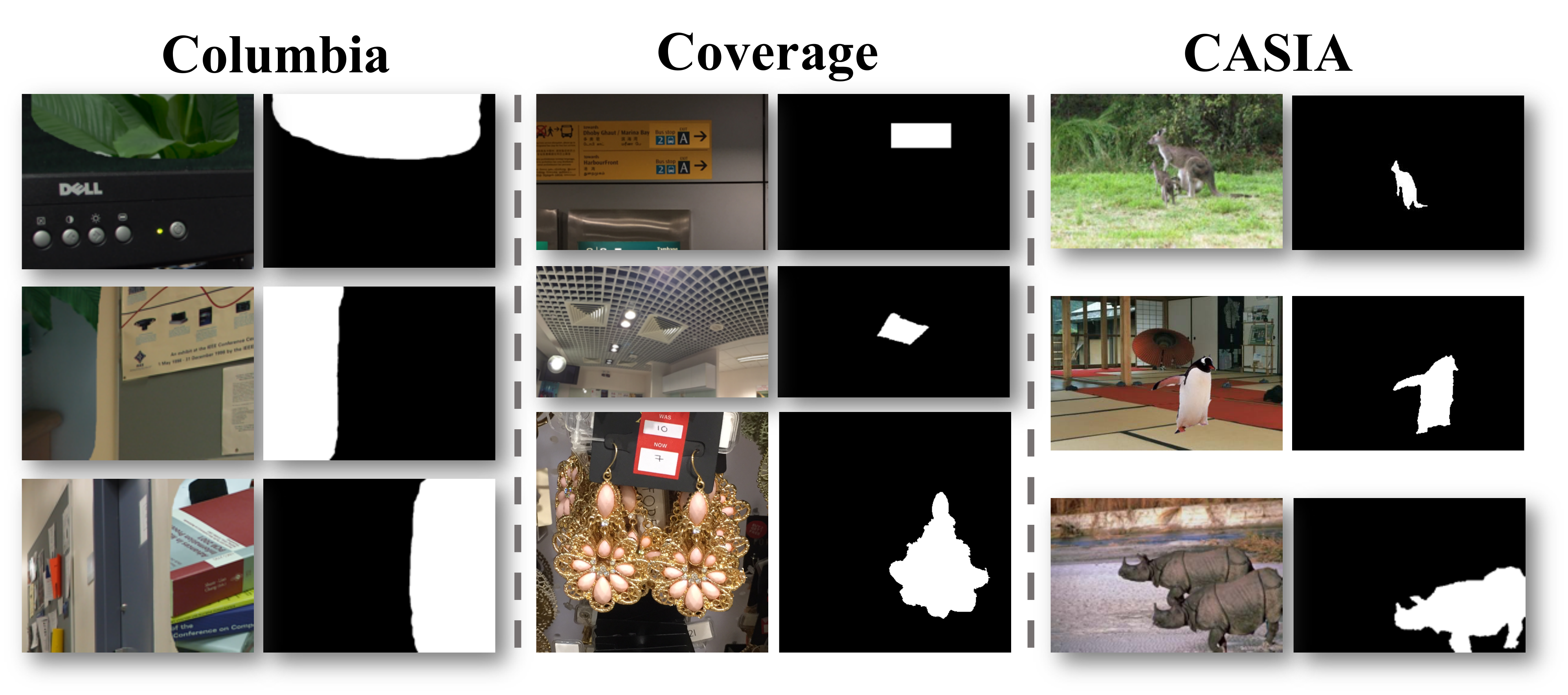}
        \caption{The characteristic difference between Columbia, Coverage and CASIA.}
        \label{fig:my_label}
    \end{figure}

    In Section 4.3 we mentioned that the characteristic difference between Columbia, Coverage and CASIA could be a possible reason of the different scale of performance improvement the SPAN achieved over RGB-N\cite{rgbn}. In Figure \ref{fig:my_label}, three images from each of these three datasets are presented. As shown, in CASIA, the manipulated objects tend to correspond to semantic objects compared to the other two. 
    
\section*{NIST16 Image Overlap between Traning and Testing}
    \begin{figure}
        \centering
        \includegraphics[width=\linewidth]{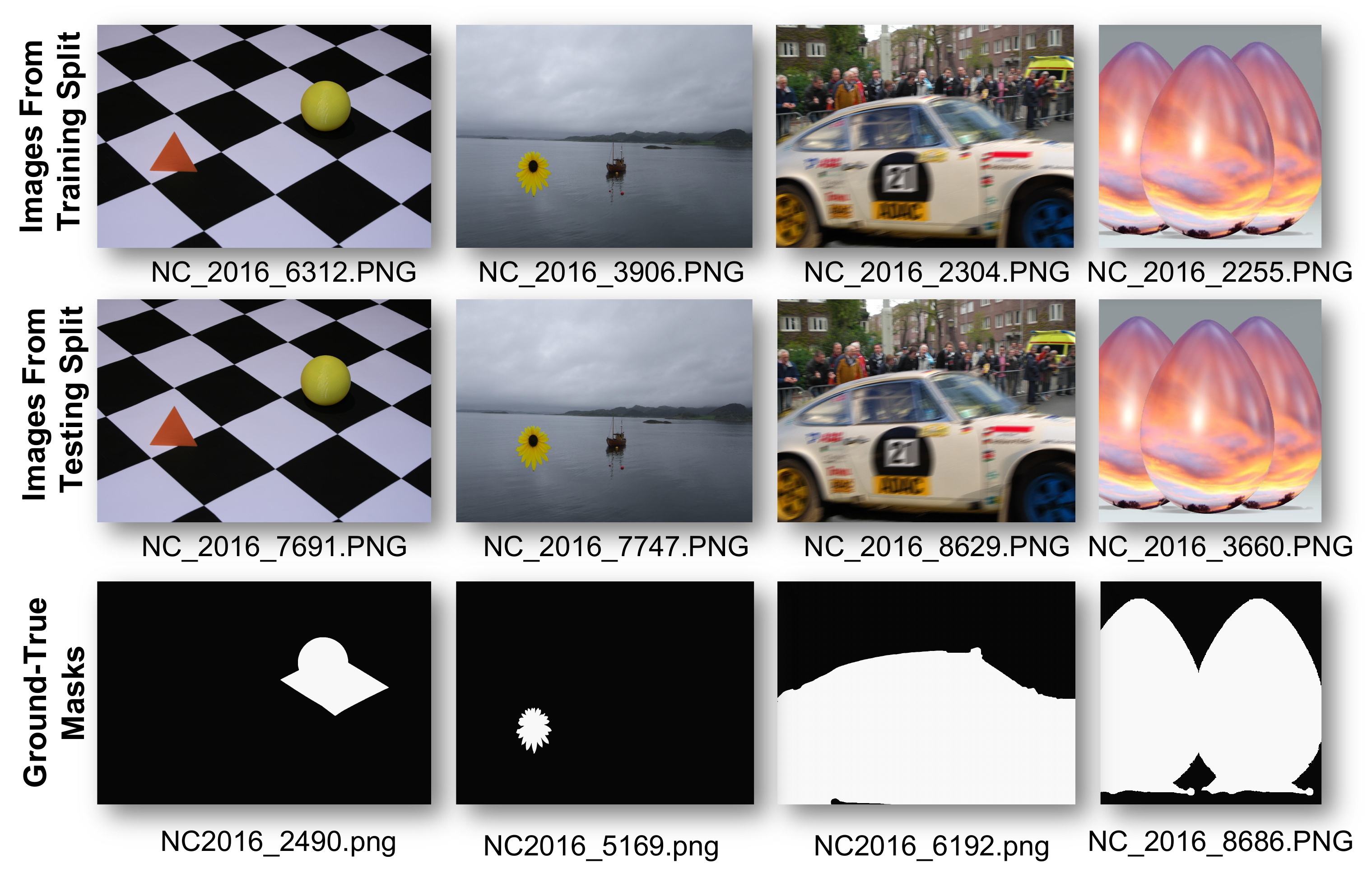}
        \caption{Demonstration of NIST16\cite{NIST16} information overlapping between its training and testing splits.}
        \label{fig:nist}
    \end{figure}
    In Section 4.3 we mentioned that there is an image overlap issue between training and test data on NIST16\cite{NIST16}, according to the training and testing split provided by RGB-N\cite{rgbn}. There are 584 images in NIST16 dataset, with 292 pairs of unique image and compressed JPEG version. Following the same practice in \cite{rgbn}, the test split contains 160 images. 151 out of the 160 (88\%) test images have the other visually similar version in training split and they share the same ground truth mask. Some overlapping images are shown in Figure \ref{fig:nist}. From top to bottom row, the image in training split, image in testing split and the ground truth mask of these two images are presented.

\section*{More Qualitative Results}
\begin{figure} [t]
        \centering
        \includegraphics[width=0.7\textwidth]{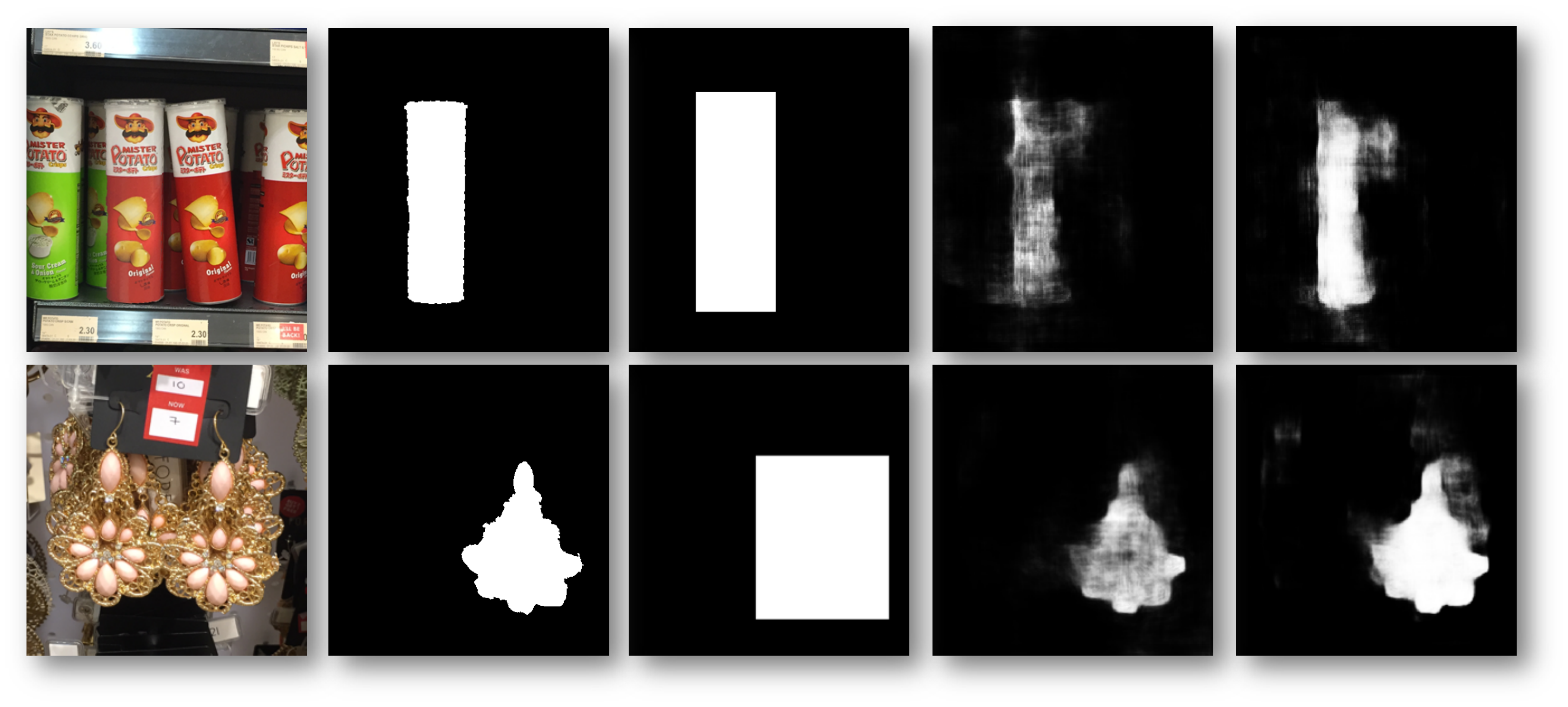}
             \caption{Qualitative Results of SPAN predictions on Coverage and NIST16 datasets, with comparison to RGB-N predictions. From left to right: Manipulated Image, Ground-truth mask, RGB-N prediction, SPAN (pre-training) prediction and SPAN(fine-tuned) prediction.}
        \label{fig:visual2}
    \end{figure}

In Figure \ref{fig:visual2}, we present the qualitative results with comparison with RGB-N\cite{rgbn}. As mentioned in Section 4.3 of the paper, RGB-N (which is built off Faster R-CNN \cite{ren2015faster}) generates region-level prediction thus only predicts rectangle mask, restricting it from generating more accurate shape of tampered region, as shown in the second row. On the other hand, there are also many test samples where the tampered region is naturally a rectangle, like the one in the first row of Figure \ref{fig:visual2}

\begin{figure}
        \centering
        \includegraphics[width=\linewidth]{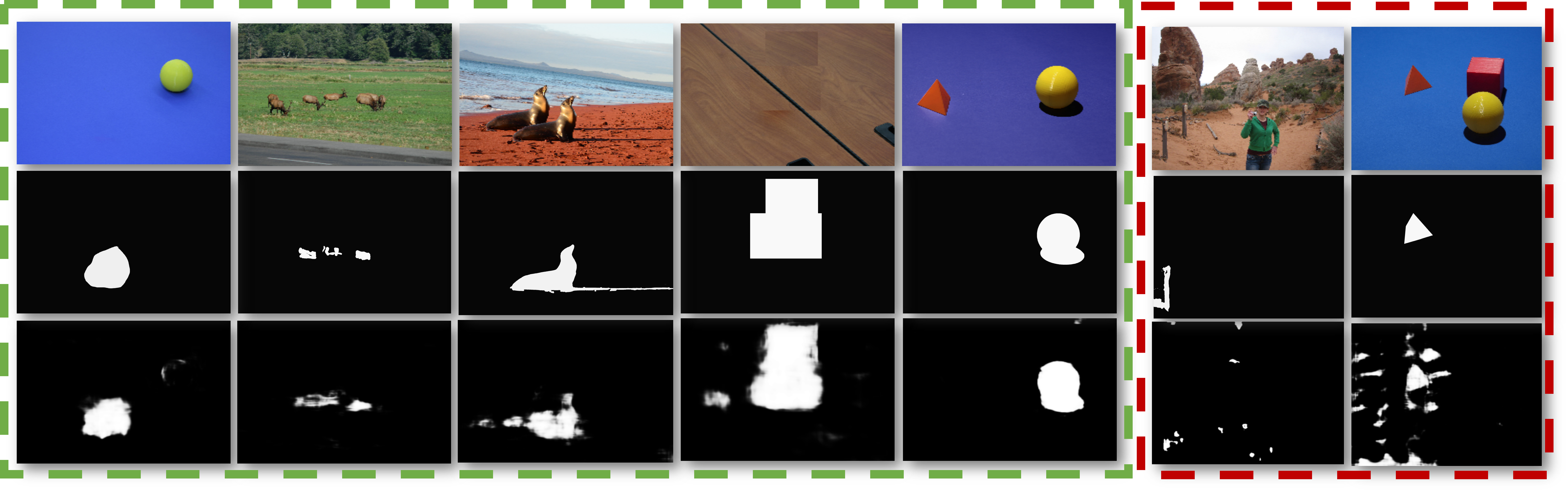}
        \caption{Qualitative Results on NIST16\cite{NIST16}. Top to bottom: Manipulated Image, Ground-Truth Mask, SPAN(fine-tuned) Prediction Mask. Red box contains failure samples.}
        \label{fig:nist_vis}
    \end{figure}

\begin{figure}
        \centering
        \includegraphics[width=\linewidth]{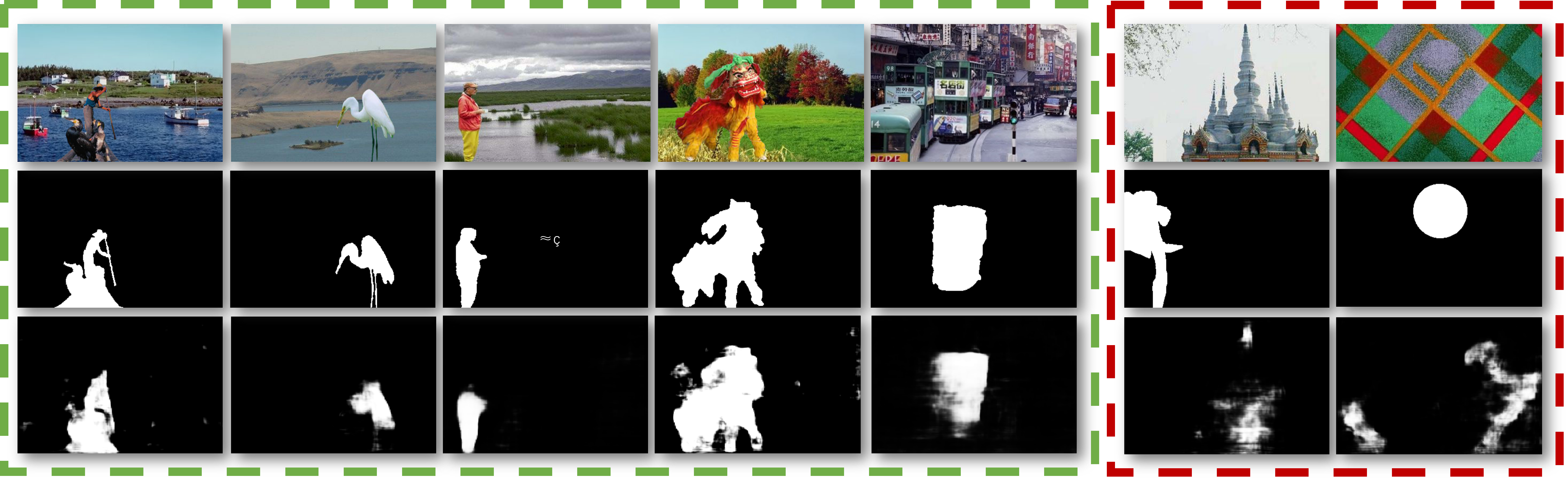}
        \caption{Qualitative Results on CASIAv1\cite{dong2013casia}. Top to bottom: Manipulated Image, Ground-Truth Mask, SPAN(fine-tuned) Prediction Mask. Red box contains failure samples.}
        \label{fig:casia_vis}
    \end{figure}

\begin{figure}
        \centering
        \includegraphics[width=\linewidth]{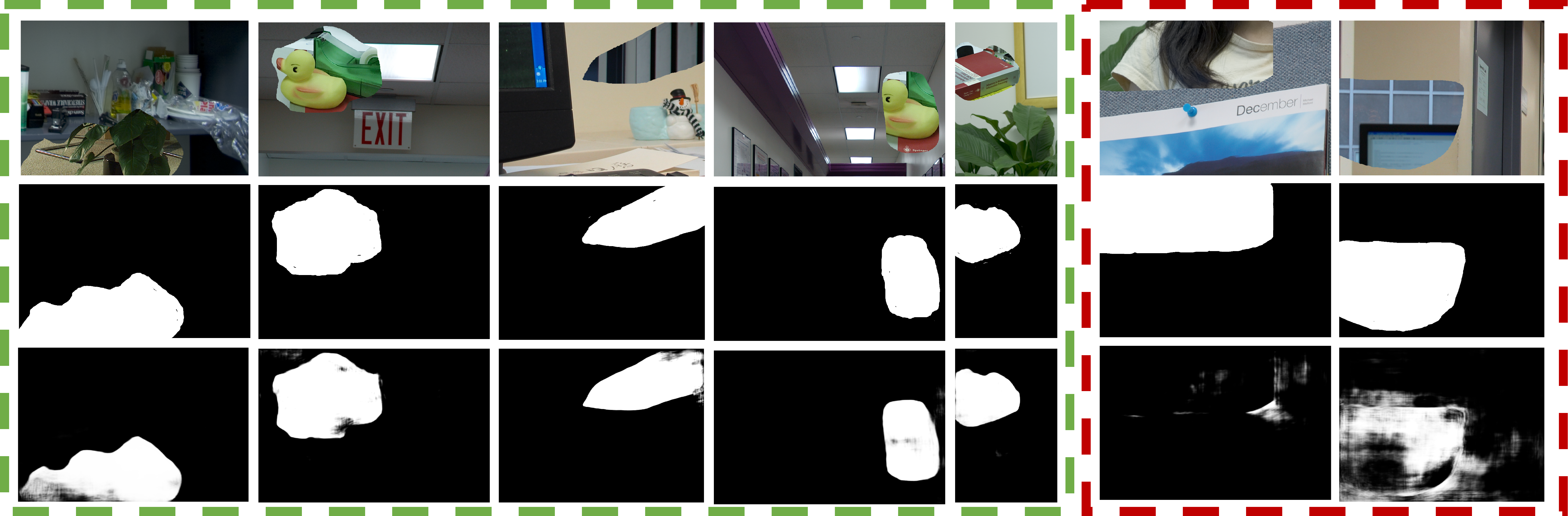}
        \caption{Qualitative Results on Columbia\cite{ng2009columbia}. Top to bottom: Manipulated Image, Ground-Truth Mask, SPAN Prediction Mask. Red box contains failure samples.}
        \label{fig:columbia_vis}
    \end{figure}

\begin{figure}
        \centering
        \includegraphics[width=\linewidth]{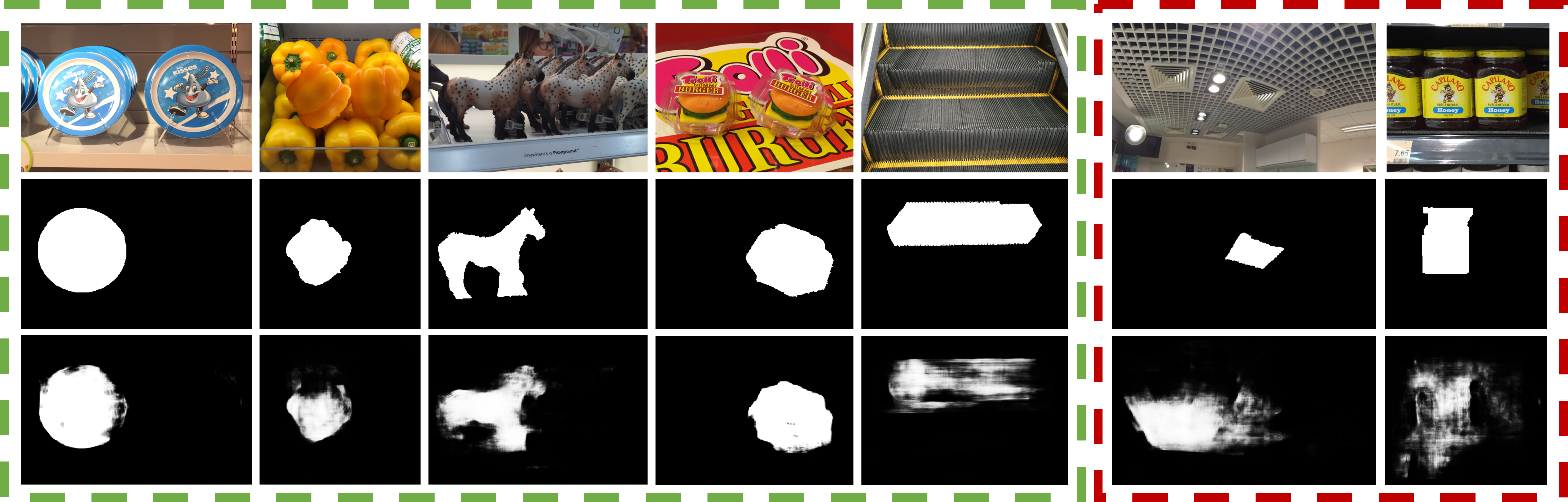}
        \caption{Qualitative Results on Coverage\cite{wen2016coverage}. Top to bottom: Manipulated Image, Ground-Truth Mask, SPAN(fine-tuned) Prediction Mask. Red box contains failure samples.}
        \label{fig:cover_vis}
    \end{figure}

We also present more visualization results of SPAN on four different datasets: NIST16 in Figure \ref{fig:nist_vis}, CASIA in \Figure \ref{fig:casia_vis}, Columbia in Figure \ref{fig:columbia_vis} and Coverage in Figure \ref{fig:cover_vis}. We sampled 5 examples (shown in first 5 columns in each image, surrounded by the green box) and 2 additional failure cases (shown in last two columns in each image, surrounded by the red box) from each dataset.

\clearpage

\bibliographystyle{splncs04}
\bibliography{egbib}